\documentclass[preprint,12pt]{elsarticle}

\usepackage{amssymb}
\usepackage{amsmath}
\usepackage{adjustbox}
\usepackage{booktabs}
\usepackage{multirow}
\usepackage{array}
\usepackage{tabularx}
\usepackage{makecell}

\usepackage{xcolor}
\usepackage{adjustbox}

\usepackage{placeins}
\usepackage{flafter}
\definecolor{incRed}{RGB}{190,40,40}
\definecolor{decGreen}{RGB}{30,130,70}
\definecolor{tieGray}{gray}{0.45}

\definecolor{GoodGreen}{RGB}{0,115,60}
\definecolor{BadRed}{RGB}{180,20,20}
\definecolor{CueBlue}{RGB}{0,75,160}

\definecolor{CorrectGreen}{RGB}{0,110,55}
\definecolor{WrongRed}{RGB}{180,20,20}

\newcommand{\bestchg}[2]{%
	\textbf{#1}\,\textcolor{incRed}{\scriptsize(#2)}%
}

\newcommand{\valchg}[2]{%
	#1\,\textcolor{decGreen}{\scriptsize(#2)}%
}

\newcommand{\good}[1]{%
	{\color{GoodGreen}#1}%
}

\newcommand{\bad}[1]{%
	{\color{BadRed}#1}%
}

\newcommand{\cue}[1]{%
	\textcolor{blue!55!black}{#1}%
}

\newcolumntype{C}[1]{%
	>{\centering\arraybackslash}m{#1}%
}

\newcolumntype{Y}{%
	>{\raggedright\arraybackslash}X%
}

\newcolumntype{L}[1]{%
	>{\raggedright\arraybackslash}p{#1}%
}

\newcommand{\QualImg}[1]{%
	\IfFileExists{#1}{%
		\parbox[c][0.070\textwidth][c]{0.115\textwidth}{%
			\centering
			\includegraphics[
			width=0.090\textwidth,
			height=0.060\textwidth
			]{#1}%
		}%
	}{%
		\fbox{%
			\parbox[c][0.060\textwidth][c]{0.090\textwidth}{%
				\centering
				\scriptsize Missing\\image%
			}%
		}%
	}%
}

\journal{Neural Networks}

\journal{Nuclear Physics B}

\begin{document}




\begin{frontmatter}
	
	\title{From Neural Tokens to Language: Dual-View EEG Semantic Bridging for Brain-to-Text Generation}
	
	\author[aff1]{YaoYang Liu}
	\ead{liuyaoyang1@mails.gdut.edu.cn}
	
	\author[aff1]{Jun Dai}
	\ead{daijun@mails.gdut.edu.cn}
	
	\author[aff1]{ZhuoShi Tang}
	\ead{tangchaoshi@mails.gdut.edu.cn}
	
	\author[aff1]{Weijun Sun\corref{cor1}}
	\ead{gdutswj@gdut.edu.cn}
	
	\author[aff1]{Xiaozhao Fang}
	\ead{xzhfang168@126.com}
	
	\author[aff2]{Jie Wen}
	\ead{jiewen_pr@126.com}
	
	\affiliation[aff1]{
		organization={School of Automation, Guangdong University of Technology},
		city={Guangzhou},
		postcode={510006},
		country={China}
	}
	
	\affiliation[aff2]{
		organization={School of Computer Science and Technology, Harbin Institute of Technology},
		city={Shenzhen},
		postcode={518055},
		country={China}
	}
	
	\cortext[cor1]{Corresponding author.}
	
	\begin{abstract}
	Visually evoked EEG-to-text generation aims to produce natural-language descriptions from neural responses elicited by visual stimuli. Existing approaches typically compress EEG signals into a single global representation and directly project it into the hidden space of a large language model (LLM), which may limit the preservation of complementary visual semantics under noisy and subject-dependent EEG signals. We propose a dual-view semantic token bridging framework that exploits raw images and sketch images as complementary visual supervision to learn raw-oriented and sketch-oriented EEG semantic token streams. To interface these structured neural representations with an LLM, we introduce a branch-specific semantic query bridge that aggregates each token stream into compact semantic slots and converts them into LLM-compatible soft prompts. A progressive adaptation strategy further transitions the learned visual-semantic interface from image-supported training to image-free EEG-conditioned generation. Experiments demonstrate strong subject-wise EEG classification performance and competitive EEG-to-text generation across multiple LLM backbones. Ablation, branch-level representation, and EEG–label disentanglement analyses further show that the two EEG streams learn complementary view-oriented representations and that EEG-derived prompts provide sample-specific information beyond category labels.
	\end{abstract}
	
	
\end{frontmatter}



\section{Introduction}
Decoding visually evoked electroencephalography (EEG) signals aims to recover perceptual information from neural responses to visual stimuli, with applications in brain--computer interfaces, computational neuroscience, and cross-modal understanding~\cite{deshpande2017new,orban2022review,chaddad2023electroencephalography,craik2019deep,roy2019deep}. Recent advances in multimodal neural decoding have extended EEG-based
visual decoding from closed-set classification and retrieval toward
semantic alignment and image reconstruction
\cite{song2024decoding,li2024visual,fu2025brainvis,
	xu2026neurovista,cao2026eegclip,cheng2027finegrained,
	xu2027umind}.
More recently, large language models (LLMs) have further enabled
open-ended neural-to-language generation
\cite{mishra2025thought2text}. Visually evoked EEG-to-text generation is particularly challenging because it must recover not only object categories but also fine-grained attributes, including shape, color, texture, and scene context.
\begin{figure}[t]
	\centering
	\includegraphics[width=0.9\columnwidth,trim=3mm 6mm 3mm 3mm,clip]{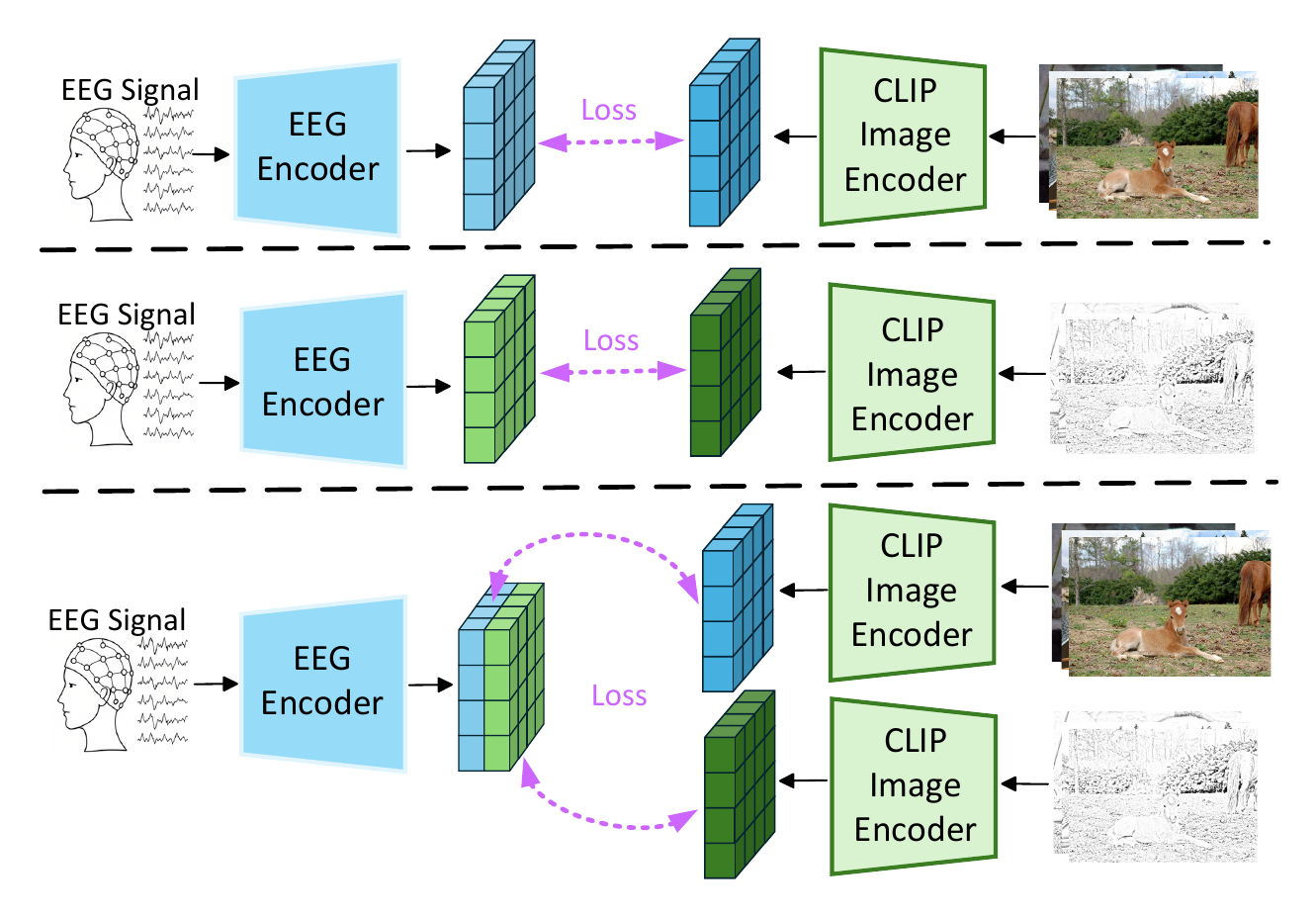}
	\caption{
		Motivation of dual-view EEG supervision.
		Instead of aligning EEG features with a single visual space, our method uses raw and sketch views as complementary supervision to learn complementary raw-oriented and sketch-oriented EEG
		representations.
	}
	\label{fig:innovation}
\end{figure}
Existing methods commonly encode EEG signals into a single global representation and align it with visual embeddings, textual embeddings, or an LLM hidden space~\cite{radford2021learning,du2023decoding,shi2025brainalign,chen2024visual,mishra2025thought2text}. Although these approaches demonstrate the semantic decodability of visually evoked EEG, they face three limitations. First, the noisy, trial-variable, and subject-dependent EEG signals are difficult to compress into a stable global representation~\cite{chaddad2023electroencephalography,craik2019deep,roy2019deep,gifford2022large}. Second, the single-pathway supervision may not adequately preserve both relatively stable structural cues, such as category, contour, and shape, and more variable appearance cues, such as color, texture, and local details. Third, directly mapping EEG features into an LLM hidden space overlooks the distributional mismatch among neural, visual, and linguistic representations, potentially causing category errors, missing details, and generic descriptions.

We therefore propose a dual-view semantic token bridging framework for visually evoked EEG-to-text generation. Rather than representing EEG with a single global vector, our framework learns complementary token streams under raw-image and sketch-image supervision. Raw images and sketches provide complementary visual cues,
particularly in appearance and structural information
~\cite{luo2023review,lei2023emergent}. Accordingly, our branch-specific supervision is
designed to induce complementary raw-oriented and
sketch-oriented EEG semantic representations. We further introduce a branch-specific semantic query bridge that aggregates the dual-view EEG tokens into compact semantic slots and converts them into LLM-compatible soft prompts. This structured interface preserves the dual-view organization
of EEG semantics during neural-to-language transfer rather than
collapsing them through a single direct projection. Extensive experiments across EEG encoders, LLM backbones, and inference-time interventions validate the proposed framework, with
additional visual-backbone results provided in the supplementary material.

Our contributions are summarized as follows:
\begin{itemize}
	\item We introduce dual-view EEG semantic tokenization that
	uses complementary raw-image and sketch-image supervision to
	induce view-specific specialization within a structured EEG
	representation.
	
	\item We introduce a branch-specific semantic query bridge that
	preserves this dual-view structure during EEG-to-LLM transfer,
	independently aggregating the two EEG token streams into compact
	semantic slots before constructing LLM-compatible soft prompts.
	
	\item Extensive experiments across EEG, vision, and LLM
	backbones, together with branch-level representation analysis
	and EEG--label disentanglement, demonstrate complementary
	view-oriented representations and sample-specific EEG information
	beyond category labels.
\end{itemize}

\section{Related Work}

\subsection{Visually Evoked Neural Decoding}
\label{sec:related_neural_decoding}
Visually evoked neural decoding seeks to recover perceptual information from brain responses elicited by visual stimuli~\cite{spampinato2017deep,gifford2022large,schirrmeister2017deep,lawhern2018eegnet}. Research has progressed from closed-set category recognition to image retrieval, semantic matching, reconstruction, and open-ended generation, indicating that EEG and MEG signals encode both category-discriminative and higher-level visual information~\cite{jiao2019decoding,du2023decoding,shi2025brainalign,chen2024visual,mishra2025thought2text}. Recent studies have further advanced EEG-based visual decoding from
category-level recognition toward richer semantic and generative tasks.
EEG-CLIP employs a Transformer-based EEG encoder and CLIP-guided
generative framework for EEG-to-image reconstruction
\cite{cao2026eegclip}, while NeuroVista introduces masked EEG modeling
and bidirectional cross-modal fusion to improve robust neural--visual
alignment \cite{xu2026neurovista}. More recently, fine-grained
multimodal EEG decoding has incorporated language-derived semantic
supervision to recover visual attributes beyond object categories
\cite{cheng2027finegrained}, and UMind jointly aligns neural,
visual, and textual representations for zero-shot visual retrieval,
classification, and reconstruction \cite{xu2027umind}. However, EEG remains noisy, trial-variant, and subject-dependent, making stable semantic representation difficult~\cite{chaddad2023electroencephalography,craik2019deep,
	roy2019deep,ng2024subjectindependent}. Many existing methods compress EEG into a single global vector, which may entangle category, shape, color, texture, and background information and thereby limit fine-grained semantic generation~\cite{du2023decoding,chen2024visual,mishra2025thought2text}. This motivates a tokenized representation that preserves complementary
view-oriented visual semantics.

Beyond EEG-based decoding, semantic priors have also been shown to facilitate the reconstruction of perceived and imagined visual content
from other neural recording modalities, such as fMRI
\cite{koidemajima2024mental}.

\subsection{Visual Semantic Alignment}
\label{sec:related_visual_alignment}
Visual semantic alignment improves EEG representation learning by mapping neural features into image or text embedding spaces constructed by pretrained visual and vision-language models~\cite{radford2021learning,jiao2019decoding,shi2025brainalign,chen2024visual}. Such alignment supports cross-modal classification, retrieval, and
semantic decoding, but many existing approaches still organize neural
representations around a unified or globally aligned semantic space
\cite{jiao2019decoding,shi2025brainalign,chen2024visual,
	mishra2025thought2text}. Raw-image embeddings retain color, texture, and background information but also contain appearance-specific variation, whereas text or global vision-language embeddings emphasize category semantics while discarding visual details. Sketch embeddings instead preserve contour and shape but suppress color and texture. Consequently, single-space alignment may bias EEG representations toward only one type of visual cue and fail to preserve both structural semantics and appearance details. Recent EEG visual-decoding approaches have increasingly strengthened
such cross-modal alignment using pretrained multimodal representations.
EEG-CLIP learns EEG representations compatible with CLIP image features
for diffusion-based reconstruction \cite{cao2026eegclip}, whereas
NeuroVista introduces bidirectional cross-modal attention to improve
fine-grained neural--visual interaction \cite{xu2026neurovista}.
Cheng et al. further align EEG signals with rich language-derived
semantic embeddings to capture attribute-level visual information
\cite{cheng2027finegrained}. UMind extends this direction by jointly
aligning neural signals with both visual and textual representations
within a unified multitask framework \cite{xu2027umind}.

\subsection{Multi-View Visual Supervision for EEG}

Multi-view learning exploits complementary information across views or
modalities and has been widely studied in recognition, retrieval, and
multimodal analysis \cite{li2016survey,zhao2017multiview,arevalo2017gmu}.
Raw images and sketches form a particularly complementary pair:
raw images preserve color, texture, material, and background, whereas
sketches emphasize contour, shape, and spatial structure.
Because EEG signals exhibit substantial noise, trial variability, and
subject dependence \cite{chaddad2023electroencephalography,
	craik2019deep,roy2019deep}, reliance on a single visual semantic view
may limit the robustness of learned representations.
Joint raw-image and sketch supervision therefore provides complementary
appearance and structural cues, reducing dependence on a single semantic
view and supporting more stable and complete EEG representations.

\subsection{LLM-Based EEG-to-Text Generation}
\label{sec:related_eeg_to_text}
Language-derived semantics have also begun to play a more explicit role
in recent EEG visual decoding. Cheng et al. employ captions generated
by a vision-language model as rich semantic supervision and align EEG
signals with text embeddings before visual reconstruction
\cite{cheng2027finegrained}. Similarly, UMind jointly exploits image
and text modalities to learn visual and semantic neural representations
for multiple downstream decoding tasks \cite{xu2027umind}.
These approaches demonstrate the value of linguistic semantics for EEG
representation learning, but their primary outputs remain visual
retrieval, classification, or reconstruction rather than direct
natural-language generation. Neural-to-text generation translates brain activity into natural-language descriptions~\cite{tang2023semantic,wang2022openvocabulary,mishra2025thought2text}. Earlier methods mapped neural signals to fixed vocabularies, semantic labels, or sentence representations before sequence generation, whereas recent neural-to-text approaches increasingly interface neural representations with pretrained language models~\cite{mishra2025thought2text,tang2023semantic,wang2022openvocabulary}. More broadly, soft-prompt, prefix-tuning, and parameter-efficient adaptation strategies have been developed
for language and vision--language models
~\cite{houlsby2019parameter,hu2021lora,sung2022vl}. Nevertheless, existing methods commonly project EEG features directly into the LLM hidden space, overlooking the distributional gaps among neural signals, visual semantic representations, and language tokens~\cite{mishra2025thought2text}. Such direct projection may cause semantic discontinuity, missing visual details, and excessive reliance on language priors. Our method instead employs branch-specific semantic querying to organize dual-view EEG representations into compact semantic slots before converting them into LLM-compatible soft prompts.

\begin{figure*}[!t]
	\centering
	\includegraphics[width=1.0\textwidth,trim=6mm 6mm 2mm 2mm,clip]{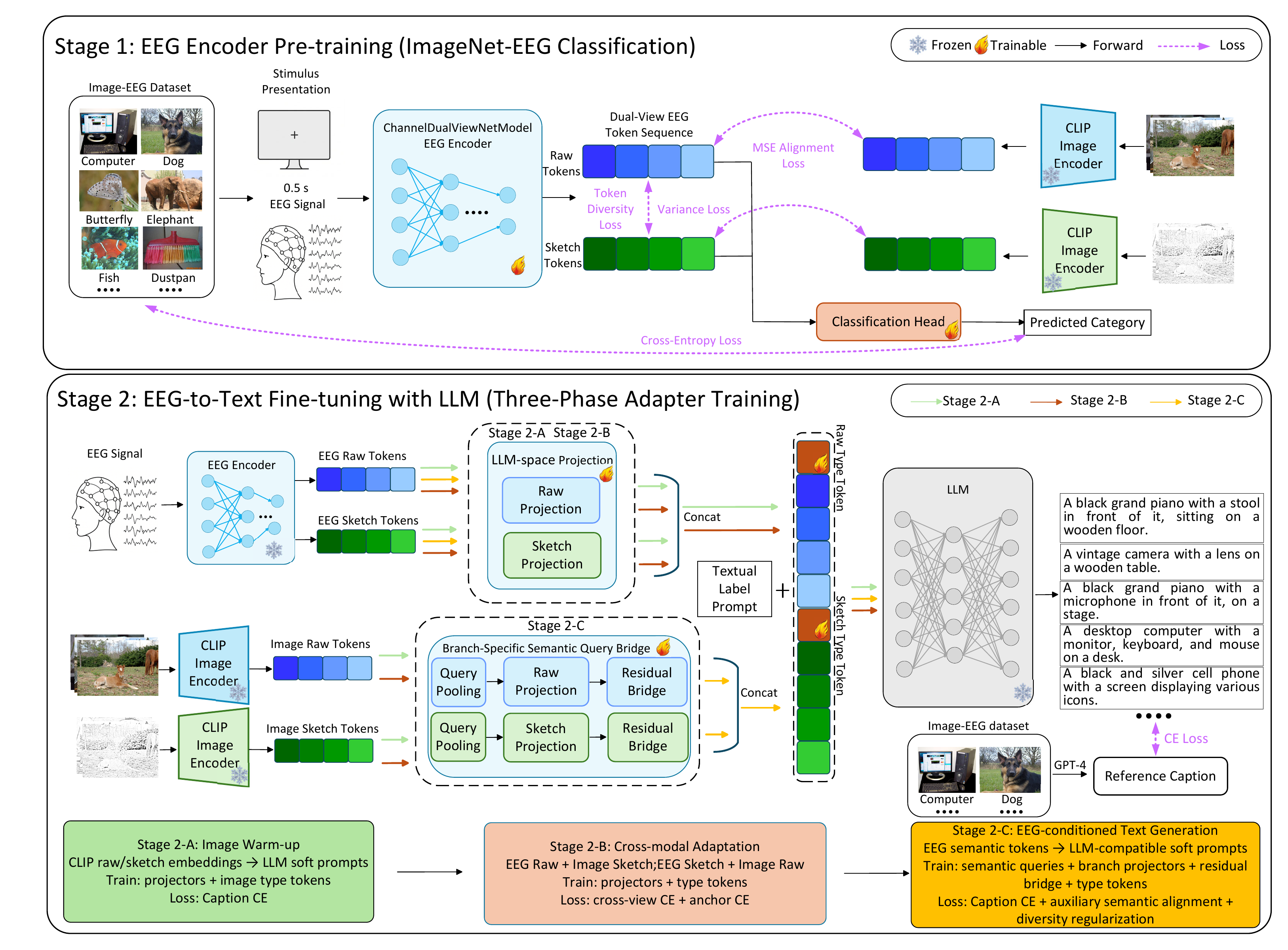}
	\caption{
		Overview of the proposed two-stage EEG-to-text framework. 
	}
	\label{fig:overall_framework}
\end{figure*}

\section{Methodology}
\label{sec:method}
Fig.~\ref{fig:overall_framework} illustrates the proposed framework,
which first learns dual-view EEG semantic tokens and then progressively
adapts them to an LLM through image warm-up, cross-modal adaptation,
and image-free EEG-conditioned semantic query adaptation.
\subsection{Dual-View EEG Semantic Token Encoder}
\label{sec:dual_view_encoder}

Given EEG inputs $\mathbf{X}\in\mathbb{R}^{B\times C_{\mathrm{in}}\times H\times W}$, a convolutional EEG backbone extracts a spatiotemporal feature map $\mathbf{F}=f_{\theta}(\mathbf{X})\in\mathbb{R}^{B\times C\times H'\times W'}$ using temporal, spatial, and residual convolutional blocks~\cite{schirrmeister2017deep,lawhern2018eegnet,song2023eegconformer}. We then pool $\mathbf{F}$ into $K$ slots and obtain the EEG token sequence by
\begin{equation}
	\begin{aligned}
		\mathbf{T}
		&=
		\mathrm{LN}
		\left(
		\operatorname{Reshape}
		\left(
		\operatorname{Pool}(\mathbf{F})
		\right)
		\mathbf{W}_{t}
		+
		\mathbf{b}_{t}
		+
		\mathbf{P}
		\right) \\
		&=
		[\mathbf{T}^{s};\mathbf{T}^{r}]
		\in
		\mathbb{R}^{B\times K\times D},
		\qquad
		K=2M,
	\end{aligned}
	\label{eq:eeg_tokenization}
\end{equation}
where $\mathbf{P}\in\mathbb{R}^{1\times K\times D}$ contains learnable positional embeddings~\cite{vaswani2017attention}, and $\mathbf{T}^{s},\mathbf{T}^{r}\in\mathbb{R}^{B\times M\times D}$ denote the sketch-view and raw-view token subsets. Their semantic roles are induced by the branch-specific visual supervision described below, rather than by the split itself.

For each branch $v\in\{s,r\}$, attention pooling aggregates the token sequence, followed by a branch-specific projection:
\begin{equation}
	\begin{aligned}
		\boldsymbol{\alpha}^{v}
		&=
		\operatorname{softmax}
		\left(
		\mathbf{T}^{v}\mathbf{w}_{a}^{v}
		+
		b_{a}^{v}
		\right), \\
		\mathbf{g}^{v}
		&=
		(\boldsymbol{\alpha}^{v})^{\top}\mathbf{T}^{v},
		\qquad
		\mathbf{z}^{v}
		=
		\phi_{v}(\mathbf{g}^{v}),
	\end{aligned}
	\label{eq:eeg_branch_encoding}
\end{equation}
where $\boldsymbol{\alpha}^{v}\in\mathbb{R}^{B\times M}$ denotes token attention weights, $\mathbf{g}^{v}\in\mathbb{R}^{B\times D}$ is the pooled branch representation, and $\phi_v$ maps it into the corresponding visual semantic space.

The two branch representations are jointly supervised by a category classifier:
\begin{equation}
	\mathcal{L}_{\mathrm{cls}}
	=
	\operatorname{CE}
	\left(
	\mathbf{W}_{c}
	(\mathbf{g}^{s}+\mathbf{g}^{r})
	+
	\mathbf{b}_{c},
	\mathbf{y}
	\right),
	\label{eq:eeg_cls_loss}
\end{equation}
where $\mathbf{y}$ denotes the ground-truth category labels. This objective supplies shared category supervision, while the view-specific alignment below differentiates the semantic roles of the two branches.

A frozen CLIP image encoder extracts sketch- and raw-view targets~\cite{radford2021learning}:
\begin{equation}
	\begin{aligned}
		\mathbf{h}^{v}
		&=
		E_{\mathrm{CLIP}}(\mathbf{I}^{v}),
		\qquad v\in\{s,r\}, \\
		\mathcal{L}_{\mathrm{align}}
		&=
		\sum_{v\in\{s,r\}}
		\lambda_v
		\mathcal{D}
		\left(
		\mathbf{z}^{v},
		\mathbf{h}^{v}
		\right),
	\end{aligned}
	\label{eq:dual_view_alignment}
\end{equation}
where $\mathbf{I}^{s}$ and $\mathbf{I}^{r}$ are the sketch and raw images, respectively, and $\mathcal{D}$ is implemented as mean squared error. Accordingly, sketch-view supervision encourages the EEG
representation toward sketch-oriented visual semantics, whereas
raw-view supervision encourages alignment with the richer
semantics preserved in the original image.
\begin{figure*}[!t]
	\centering
	\includegraphics[width=1.0\textwidth,trim=2mm 6mm 2mm 2mm,clip]{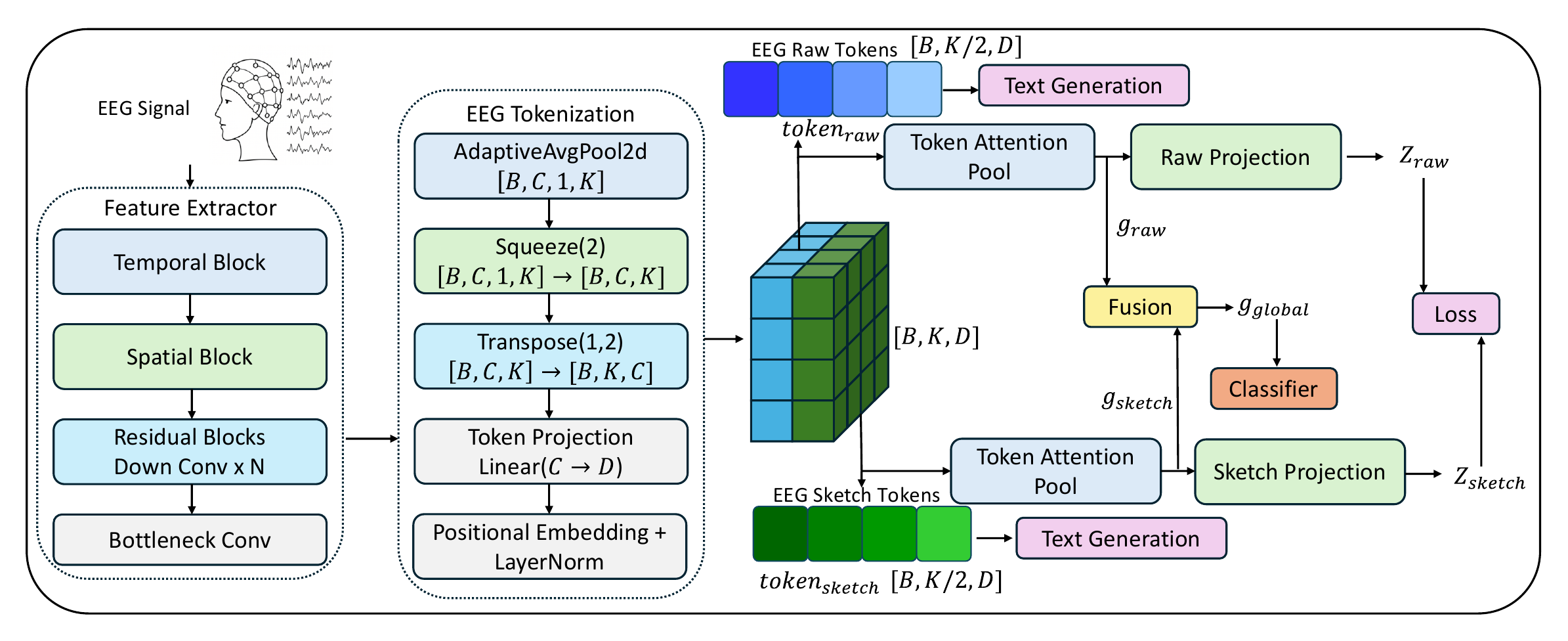}
	\caption{
		Architecture of the proposed dual-view EEG semantic token encoder.
	}
	\label{fig:eeg_backbone}
\end{figure*}
We further regularize the token space to reduce redundancy and representation collapse~\cite{zbontar2021barlow,bardes2021vicreg}. Let $\bar{\mathbf{T}}_{b}^{v}$ denote the $\ell_2$-normalized tokens of sample $b$, and let $\tilde{\mathbf{T}}\in\mathbb{R}^{BK\times D}$ be the flattened full token sequence. The diversity and variance terms are defined compactly as
\begin{equation}
	\begin{aligned}
		\mathbf{S}_{b}^{v}
		&=
		\bar{\mathbf{T}}_{b}^{v}
		(\bar{\mathbf{T}}_{b}^{v})^{\top}, \\
		\mathcal{L}_{\mathrm{div}}
		&=
		\frac{1}{2BM(M-1)}
		\sum_{v\in\{s,r\}}
		\sum_{b=1}^{B}
		\left\|
		\operatorname{OffDiag}
		(\mathbf{S}_{b}^{v})
		\right\|_{F}^{2}, \\
		\mathcal{L}_{\mathrm{var}}
		&=
		\frac{1}{D}
		\sum_{d=1}^{D}
		\max
		\left(
		0,
		\gamma-
		\sqrt{
			\operatorname{Var}
			(\tilde{\mathbf{T}}_{:,d})
			+
			\epsilon
		}
		\right),
	\end{aligned}
	\label{eq:eeg_token_regularization}
\end{equation}
where $\operatorname{OffDiag}(\cdot)$ retains only off-diagonal entries, $\gamma$ is the target standard-deviation margin, and $\epsilon$ ensures numerical stability. The diversity term discourages redundant tokens within each branch, while the variance term preserves feature-wise variation.

The Stage-1 EEG encoder is optimized using
\begin{equation}
	\mathcal{L}_{\mathrm{EEG}}
	=
	\mathcal{L}_{\mathrm{cls}}
	+
	\mathcal{L}_{\mathrm{align}}
	+
	\lambda_{\mathrm{div}}^{\mathrm{eeg}}
	\mathcal{L}_{\mathrm{div}}
	+
	\lambda_{\mathrm{var}}
	\mathcal{L}_{\mathrm{var}},
	\label{eq:eeg_encoder_total_loss}
\end{equation}
where $\lambda_{\mathrm{div}}^{\mathrm{eeg}}=\lambda_{\mathrm{var}}=0.05$. After pretraining, $\mathbf{T}^{s}$ and $\mathbf{T}^{r}$ are used as complementary neural semantic inputs for EEG-to-text generation.

\subsection{Progressive EEG-to-LLM Adaptation}
\label{sec:progressive_adapter}

The Stage-1 encoder produces complementary sketch-view and raw-view EEG tokens, $\mathbf{T}^{s}$ and $\mathbf{T}^{r}$. Because these neural representations are not naturally aligned with the LLM token space, directly projecting them into the LLM hidden space may cause semantic mismatch. We therefore progressively adapt the EEG tokens through image-based
warm-up, image--EEG cross-view adaptation, and image-free EEG-conditioned semantic query adaptation.

\subsubsection{Multimodal Soft-Prompt Construction}

Let $\mathcal{S}=\big((q_i,\mathbf{x}_i)\big)_{i=1}^{J}$ be an ordered collection of multimodal inputs, where $q_i$ identifies the modality and view of feature block $\mathbf{x}_i$. Each block is projected into the LLM hidden dimension $H_{\ell}$ and concatenated with a learnable type token:
\begin{equation}
	\begin{aligned}
		\mathbf{m}_{i}
		&=
		\Pi_{v_i}(\mathbf{x}_{i})
		\in
		\mathbb{R}^{B\times L_i\times H_{\ell}},\\
		\mathbf{E}(\mathcal{S})
		&=
		\left[
		\mathbf{E}^{\mathrm{pre}};
		\operatorname*{Concat}_{i=1}^{J}
		\left(
		\boldsymbol{\tau}_{q_i},
		\mathbf{m}_{i}
		\right);
		\mathbf{E}^{\mathrm{post}}
		\right],\\
		\mathcal{L}_{\mathrm{cap}}(\mathcal{S})
		&=
		-\frac{1}{L}
		\sum_{t=1}^{L}
		\log
		P_{\Theta}
		\left(
		y_t
		\mid
		y_{<t},
		\mathbf{E}(\mathcal{S})
		\right).
	\end{aligned}
	\label{eq:multimodal_prompt_caption}
\end{equation}
where $\Pi_{v_i}$ is the sketch- or raw-branch projector, $\boldsymbol{\tau}_{q_i}$ is the corresponding modality--view type token, and $\mathbf{E}^{\mathrm{pre}}$ and $\mathbf{E}^{\mathrm{post}}$ are the token embeddings before and after the multimodal placeholder~\cite{li2021prefix,lester2021power}. The caption loss is computed only over the target caption tokens $\mathbf{Y}=\{y_t\}_{t=1}^{L}$.

\paragraph{Prompt construction}
For EEG-conditioned text generation, we use a fixed chat-style
instruction across training and evaluation. The multimodal placeholder
denotes the insertion position of the projected multimodal or EEG soft
prompt rather than an actual image input during EEG-only inference.
The object category is provided as an auxiliary textual label prompt.
During training, ground-truth category names are used and the reference
caption serves as the autoregressive generation target. During Stage C
inference, the category label is obtained from the frozen Stage-1 EEG
classifier when label prompting is enabled. The exact prompt templates
for the evaluated LLM interfaces are provided in the Supplementary
Material.

\subsubsection{Stage A: Dual-View Image Warm-Up}

Stage A initializes the visual-to-language interface using the more stable image semantics before introducing noisy EEG features~\cite{li2023blip2}. We reuse the frozen CLIP embeddings
$\mathbf{h}^{v}=E_{\mathrm{CLIP}}(\mathbf{I}^{v})$ for $v\in\{s,r\}$ and define
\begin{equation}
	\begin{aligned}
		\mathcal{S}_{A}
		&=
		\big(
		(\mathrm{img}\text{-}s,\mathbf{h}^{s}),
		(\mathrm{img}\text{-}r,\mathbf{h}^{r})
		\big),\\
		\mathcal{S}_{A}^{v}
		&=
		\big(
		(\mathrm{img}\text{-}v,\mathbf{h}^{v})
		\big),\\
		\mathcal{L}_{A}
		&=
		\mathcal{L}_{\mathrm{cap}}(\mathcal{S}_{A})
		+
		\frac{\lambda_{\mathrm{single}}}{2}
		\sum_{v\in\{s,r\}}
		\mathcal{L}_{\mathrm{cap}}(\mathcal{S}_{A}^{v}).
	\end{aligned}
	\label{eq:stage_a_objective}
\end{equation}
The joint prompt learns complementary raw/sketch conditioning, while the single-view terms stabilize the two branch projectors. The EEG encoder and LLM backbone remain frozen, and only the multimodal interface parameters are updated.

\subsubsection{Stage B: Image-Anchored Cross-Modal Adaptation}

Stage B gradually introduces EEG tokens by pairing each neural branch with the complementary image view:
\begin{equation}
	\begin{aligned}
		\mathcal{S}_{B}^{1}
		&=
		\big(
		(\mathrm{eeg}\text{-}r,\mathbf{T}^{r}),
		(\mathrm{img}\text{-}s,\mathbf{h}^{s})
		\big),\\
		\mathcal{S}_{B}^{2}
		&=
		\big(
		(\mathrm{eeg}\text{-}s,\mathbf{T}^{s}),
		(\mathrm{img}\text{-}r,\mathbf{h}^{r})
		\big),\\
		\mathcal{L}_{B}
		&=
		\frac{1}{2}
		\sum_{k=1}^{2}
		\mathcal{L}_{\mathrm{cap}}(\mathcal{S}_{B}^{k})
		+
		\lambda_{\mathrm{anchor}}
		\mathcal{L}_{\mathrm{cap}}(\mathcal{S}_{A}).
	\end{aligned}
	\label{eq:stage_b_objective}
\end{equation}
where $\lambda_{\mathrm{single}}$ and $\lambda_{\mathrm{anchor}}$ weight the
single-view and image-anchor objectives, respectively. The cross-view pairs encourage interaction between the
complementary raw-oriented and sketch-oriented EEG semantics,
whereas the image-only anchor preserves the visual-to-language
interface learned in Stage A. As in
Stage A, the EEG encoder and LLM backbone remain frozen while the
multimodal interface is optimized. During both Stages A and B, the
ground-truth category name is inserted into the textual prompt, and the
corresponding reference caption serves as the autoregressive generation
target; predicted category labels are not used during training.

\subsubsection{Stage C: Branch-Specific Semantic Query Bridging}
Stage C continues from the Stage-B model state and removes all visual
inputs for image-free EEG-conditioned generation. It retains and further
optimizes the learned EEG modality--view type tokens while keeping the
Stage-1 EEG encoder and LLM backbone frozen. The frozen EEG encoder
produces sketch-view and raw-view token sequences
$\mathbf{T}^{s},\mathbf{T}^{r}\in\mathbb{R}^{B\times M\times D}$.
Rather than directly projecting these tokens into the LLM input space,
we employ branch-specific learnable queries to extract compact semantic
slots, which are subsequently mapped to and aligned with semantic
representations derived from the frozen LLM token-embedding layer.
Accordingly, only the branch-specific queries, EEG-to-text projectors,
residual bridges, and EEG type tokens are optimized in this stage.

For each branch $v\in\{s,r\}$, $N$ learnable queries aggregate the
$M$ EEG tokens as
\begin{equation}
	\begin{aligned}
		\mathbf{G}^{v}
		&=
		\operatorname{softmax}
		\left(
		\frac{
			\mathbf{Q}^{v}(\mathbf{T}^{v})^{\top}
		}{
			\sqrt{D}
		}
		\right)
		\mathbf{T}^{v},\\
		\mathbf{Z}^{v}
		&=
		\Psi_{v}(\mathbf{G}^{v}),
		\qquad v\in\{s,r\},
	\end{aligned}
	\label{eq:eeg_text_slot_projection}
\end{equation}
where $\mathbf{Q}^{v}\in\mathbb{R}^{N\times D}$ contains the
learnable queries, $\mathbf{G}^{v}\in\mathbb{R}^{B\times N\times D}$
denotes the resulting EEG semantic slots, and $\Psi_v$ maps them into
$\mathbf{Z}^{v}\in\mathbb{R}^{B\times N\times H_{\ell}}$, with
$H_{\ell}$ denoting the LLM hidden dimension.

A lightweight gated residual bridge converts the EEG-derived text
representations into LLM-readable soft prompts:
\begin{equation}
	\begin{aligned}
		\mathbf{R}^{v}
		&=
		\mathbf{Z}^{v}
		+
		\eta_v
		\operatorname{MLP}_{v}
		\left(
		\operatorname{LN}(\mathbf{Z}^{v})
		\right),\\
		\mathbf{M}_{C}
		&=
		[
		\boldsymbol{\tau}_{\mathrm{eeg}\text{-}s};
		\mathbf{R}^{s};
		\boldsymbol{\tau}_{\mathrm{eeg}\text{-}r};
		\mathbf{R}^{r}
		],
	\end{aligned}
	\label{eq:eeg_text_soft_prompt}
\end{equation}
where $\eta_v$ is a learnable gate and
$\boldsymbol{\tau}_{\mathrm{eeg}\text{-}s}$ and
$\boldsymbol{\tau}_{\mathrm{eeg}\text{-}r}$ are learnable branch-type
tokens. Since $\mathbf{R}^{v}$ is already represented in the LLM
hidden dimension, no additional multimodal projection is applied.
The resulting soft prompt is concatenated with the textual prefix and
suffix embeddings. Stage C training uses the ground-truth category name as the textual label prompt, whereas inference uses the category predicted by the frozen Stage-1 EEG classifier. Ground-truth category prompts at inference are used only for oracle analysis. 

\subsubsection{Branch Text-Semantic Alignment}

The sketch branch is supervised using frozen LLM embeddings of category
names, whereas the raw branch is supervised using frozen embeddings of
reference captions. Let
\begin{equation}
	\begin{aligned}
		\bar{\mathbf{Z}}^{v}
		&=
		\frac{1}{N}
		\sum_{n=1}^{N}
		\mathbf{Z}^{v}_{:,n},\\
		\bar{\mathbf{Y}}^{s}
		&=
		\operatorname{Mean}
		\left(
		\mathcal{E}_{\mathrm{LLM}}
		(
		\operatorname{Tok}(\mathbf{c})
		)
		\right),\\
		\bar{\mathbf{Y}}^{r}
		&=
		\operatorname{Mean}
		\left(
		\mathcal{E}_{\mathrm{LLM}}
		(
		\mathbf{y}_{\mathrm{id}}
		)
		\right),
	\end{aligned}
	\label{eq:branch_text_targets}
\end{equation}
where $\mathbf{c}$ denotes the category name,
$\mathbf{y}_{\mathrm{id}}$ denotes the reference-caption token IDs,
and $\mathcal{E}_{\mathrm{LLM}}$ is the frozen LLM embedding layer.
All text targets are detached from gradient computation.

For each branch, the alignment objective combines paired cosine
alignment and symmetric batch contrastive learning:
\begin{equation}
	\begin{aligned}
		\mathcal{L}_{\mathrm{cos}}^{v}
		&=
		1-
		\frac{1}{B}
		\sum_{b=1}^{B}
		\cos
		\left(
		\bar{\mathbf{Z}}_{b}^{v},
		\bar{\mathbf{Y}}_{b}^{v}
		\right),\\
		\mathcal{L}_{\mathrm{text}}^{v}
		&=
		\mathcal{L}_{\mathrm{cos}}^{v}
		+
		\beta_{\mathrm{con}}
		\mathcal{L}_{\mathrm{con}}^{v},\\
		\mathcal{L}_{\mathrm{text}}
		&=
		\frac{1}{2}
		\sum_{v\in\{s,r\}}
		\mathcal{L}_{\mathrm{text}}^{v}.
	\end{aligned}
	\label{eq:branch_text_alignment}
\end{equation}
For the sketch branch, EEG samples sharing the same category are treated
as multiple positives, while the raw branch uses paired EEG--caption
samples as instance-level positives. We set
$\beta_{\mathrm{con}}=0.5$ and the contrastive temperature to
$\tau=0.07$.

We additionally discourage within-branch token redundancy:
\begin{equation}
	\mathcal{L}_{\mathrm{div}}^{\mathrm{text}}
	=
	\frac{1}{2}
	\left[
	\Omega(\mathbf{Z}^{s})
	+
	\Omega(\mathbf{Z}^{r})
	\right],
	\label{eq:text_token_diversity}
\end{equation}
where $\Omega(\cdot)$ penalizes off-diagonal token similarities.
The final Stage-C objective is
\begin{equation}
	\mathcal{L}_{C}
	=
	\mathcal{L}_{\mathrm{cap}}(\mathbf{M}_{C})
	+
	\lambda_{\mathrm{text}}
	\mathcal{L}_{\mathrm{text}}
	+
	\lambda_{\mathrm{div}}^{\mathrm{text}}
	\mathcal{L}_{\mathrm{div}}^{\mathrm{text}},
	\label{eq:stage_c_objective}
\end{equation}
where $\lambda_{\mathrm{text}}=0.5$ and
$\lambda_{\mathrm{div}}^{\mathrm{text}}=0.05$. Here, $\mathcal{L}_{\mathrm{cap}}(\mathbf{M}_{C})$ denotes the
caption loss conditioned on the assembled Stage-C EEG soft prompt
and textual label prompt.

\section{Experiments}
\subsection{Dataset and Experimental Settings}

We evaluate on the ImageNet-EEG dataset~\cite{spampinato2017deep}, whose visual
stimuli are drawn from ImageNet~\cite{deng2009imagenet}. Following Thought2Text~\cite{mishra2025thought2text},
we use the GPT-4-generated single-sentence captions provided by
that work as textual supervision and construct sketch views using
Gaussian smoothing followed by Canny edge detection~\cite{mishra2025thought2text,canny1986computational}. We adopt a
predefined image-disjoint split.

EEG trials with original length $T$ satisfying
$460 \leq T \leq 600$ samples are cropped to the zero-based
interval $[20,460)$, yielding inputs with 128 channels and
440 temporal samples. The final split contains 7,959 training,
1,994 validation, and 1,987 test samples.

Unless otherwise specified, we use a frozen CLIP ViT-B/32 visual
encoder. Ablation experiments use Mistral-7B-v0.3 with ViT-B/32.
Experiments were conducted on a Linux server equipped with four
NVIDIA A30 GPUs. Stage-1 training required approximately 7 hours,
and Stage-2 fine-tuning required approximately 12 hours, using
Python 3.10.18, PyTorch 2.4.1, and CUDA 12.1.

\textbf{Evaluation Metrics.}
We evaluate EEG-to-text generation using lexical overlap,
semantic similarity, and LLM-based metrics. We report ROUGE,
BLEU, METEOR, and BERTScore. In addition, we employ GPT-4o
as an automatic language evaluator for two complementary
criteria: Fluency and Adequacy. For each test sample, the
evaluator receives the reference caption and the generated
caption. Fluency assesses grammatical naturalness and linguistic
quality independently of the reference, whereas Adequacy
evaluates the extent to which the generated caption preserves
the semantic content expressed in the reference caption.

Both criteria are rated on a 1--5 Likert scale, where higher scores
indicate better quality. The same evaluation prompt is used for all
methods and test samples, and the reported scores are averaged over
the complete test set.
\begin{table}[!htbp]
	\centering
	\caption{Classification accuracy (\%) on the ImageNet-EEG visual classification dataset~\cite{spampinato2017deep}. Best and second-best results are highlighted for directly
		comparable subject-wise columns. ``Overall'' values are retained
		for reference because published methods may follow different
		aggregate evaluation protocols.}
	\label{tab:imagenet_eeg_cls_comparison}
	\renewcommand{\arraystretch}{1.12}
	\setlength{\tabcolsep}{5.0pt}
	\resizebox{\textwidth}{!}{
		\begin{tabular}{llcccccccc}
			\toprule
			Method  & Subject1 & Subject2 & Subject3 & Subject4 & Subject5 & Subject6 & Overall & Subject Avg. \\
			\midrule
			
			Spampinato et al. re-eval.~\cite{spampinato2017deep,palazzo2021decoding}
			& -- & -- & -- & -- & -- & -- & 21.80 & -- \\
			
			EEGCVPR40~\cite{spampinato2017deep}
			& -- & -- & -- & -- & -- & -- & 26.00 & -- \\
			
			SyncNet~\cite{li2017targeting}
			& -- & -- & -- & -- & -- & -- & 27.00 & -- \\
			
			ShallowConvNet~\cite{schirrmeister2017deep}
			& 45.68 & 56.92 & 45.40 & 54.46 & 59.08 & 55.51 & 49.12 & 52.84 \\
			
			EEGNet~\cite{lawhern2018eegnet} 
			& 19.44 & 36.31 & 30.98 & 40.31 & 28.31 & 33.74 & 25.21 & 31.52 \\
			EEG-ChannelNet~\cite{palazzo2021decoding}
			& -- & -- & -- & -- & -- & -- & 36.00 & -- \\

			SigTanh Gated Transformer~\cite{tao2021gated}
			& -- & -- & -- & -- & -- & -- & 49.00 & -- \\
			
			EfficientNet + SVM~\cite{mishra2022eegfeature}
			& -- & -- & -- & -- & -- & -- & \underline{64.00} & -- \\
			
			GRU-Gated Transformer~\cite{tao2021gated}
			& 43.02 & \underline{70.52} & \underline{63.75} & \underline{73.65} & 56.25 & \underline{64.58} & 46.00 & \underline{61.96} \\
			
			ATCNet~\cite{altaheri2023atcnet}
			& 14.51 & 35.39 & 27.30 & 40.62 & 30.15 & 23.62 & 30.35 & 28.60 \\
			
			EEG Conformer~\cite{song2023eegconformer}
			& 32.41 & 52.31 & 43.87 & 56.00 & 40.92 & 44.79 & 52.44 & 45.05 \\
			
			CNN-BiLSTM~\cite{sharma2024eegclassification}
			& -- & -- & -- & -- & -- & -- & \textbf{71.00} & -- \\
			
			CNN-Transformer~\cite{sharma2024eegclassification}
			& -- & -- & -- & -- & -- & -- & 59.00 & -- \\
			EidetiCom~\cite{zheng2024eideticom}
			& 40.00 & 60.84 & 59.64 & 72.37 & 55.76 & 51.21 & -- & 56.64 \\
			
			Thought2Text~\cite{mishra2025thought2text} 
			& \underline{63.10} & 63.10 & 54.40 & 60.00 & \underline{59.40} & 53.10 & 53.00 & 58.85 \\
			
			\textbf{Ours}
			& \textbf{63.75} & \textbf{83.54} & \textbf{81.25} & \textbf{88.12} 
			& \textbf{73.33} & \textbf{66.46} & 58.94 & \textbf{76.08} \\
			
			\bottomrule
		\end{tabular}
	}
\end{table}

\subsection{EEG Classification Results}
\label{sec:diagnostic_classification}
\begin{figure*}[h]
	\centering
	\includegraphics[
	width=0.9\textwidth,
	trim=2mm 6mm 2mm 2mm,
	clip
	]{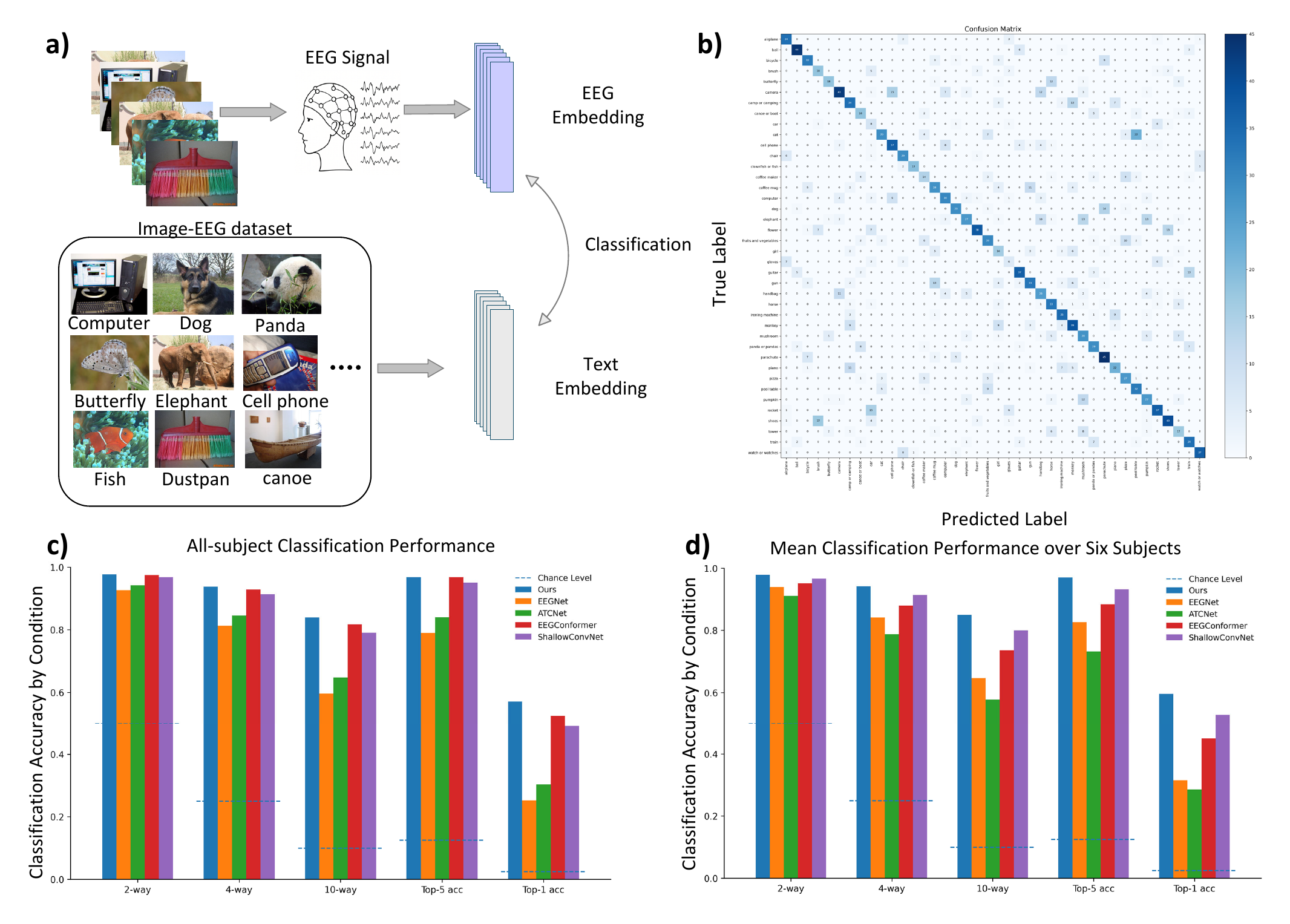}
	\caption{
		Detailed analysis of Stage-1 EEG classification.
		(a) Evaluation pipeline for category prediction from visually evoked EEG.
		(b) Confusion matrix of the proposed model on the 40-class classification
		task, showing a pronounced diagonal structure.
		(c) All-subject classification performance under 2-way, 4-way, 10-way,
		Top-5, and Top-1 evaluation settings.
		(d) Corresponding subject-wise performance averaged over the six subjects.
		Dashed horizontal lines indicate the chance level for each evaluation setting.
	}
	\label{fig:classifier_results}
\end{figure*}

To assess whether the Stage-1 EEG encoder learns category-discriminative representations, Table~\ref{tab:imagenet_eeg_cls_comparison} compares publicly reported methods, representative EEG backbones, and our model under subject-wise and all-subject settings. ``Overall'' denotes the all-subject accuracy for methods evaluated by us and the published aggregate accuracy for methods without subject-wise results, whereas ``Subject Avg.'' is the mean accuracy across the six subjects; these two columns therefore summarize different evaluation protocols and should be interpreted separately. Our model achieves a Subject Avg. accuracy of 76.08\% and
obtains the highest accuracy on all six subjects among the methods
with available subject-wise results in Table~\ref{tab:imagenet_eeg_cls_comparison}. These results indicate that complementary raw-image and sketch supervision strengthens the category-discriminative information encoded in EEG representations. All compared EEG backbones in Fig.~\ref{fig:classifier_results}
are trained and evaluated using the same data split and classification
protocol for a controlled comparison.

To further examine the category-discriminative structure learned by the
Stage-1 EEG encoder, Fig.~\ref{fig:classifier_results} provides
a more detailed classification analysis. The 40-class confusion matrix
exhibits a pronounced diagonal structure, indicating that the learned EEG
representations preserve category-specific information across a broad set
of visual classes. Consistent performance advantages are also observed
under 2-way, 4-way, 10-way, Top-5, and Top-1 evaluation settings.
Importantly, the same trend is maintained when performance is averaged
across the six subjects, suggesting that the classification gains are not
restricted to the pooled all-subject evaluation.

\subsection{EEG-to-Text Results}
\label{sec:baseline_comparison}
Table~\ref{tab:thought2text_vs_ours} compares the proposed method with Thought2Text under different LLM backbones. 
Relative to Thought2Text, our method consistently improves
ROUGE, BLEU, METEOR, BERTScore, and GPT-4o-based adequacy
across all three evaluated LLM backbones. Under Qwen2.5-7B, our method also improves ROUGE and BLEU
over SYNAPSE, while remaining lower on METEOR
(27.2 vs.~28.0) and BERTScore (0.894 vs.~0.904), indicating
that the gains over this stronger baseline are metric-dependent.
Overall, these results demonstrate that the dual-view EEG
semantic encoder and branch-specific semantic query bridge
provide an effective neural-to-language interface.
Although our method shows slightly lower GPT-4o-based fluency
in some settings, it achieves consistently higher adequacy,
indicating that the generated descriptions preserve more of the
semantic content expressed in the reference captions rather than
merely producing generic fluent sentences.

\begin{table}[!htbp]
	\centering
	\caption{
		Comparison with other methods under different LLM backbones.
		Parenthesized values report absolute changes relative to Thought2Text
		for LLaMA3-8B and Mistral-7B-v0.3, and relative to SYNAPSE for
		Qwen2.5-7B. Positive, negative, and zero values denote improvements,
		decreases, and ties, respectively.
	}
	\label{tab:thought2text_vs_ours}
	\scriptsize
	
	\setlength{\tabcolsep}{1.25pt}
	\renewcommand{\arraystretch}{1.10}
	
	\resizebox{\textwidth}{!}{%
		\begin{tabular}{@{}ll*{9}{c}@{}}
			\toprule
			
			\multirow{2}{*}{LLM}
			& \multirow{2}{*}{Method}
			& \multicolumn{3}{c}{ROUGE}
			& \multicolumn{2}{c}{BLEU}
			& \multirow{2}{*}{METEOR}
			& \multirow{2}{*}{\shortstack{BERT\\Score}}
			& \multirow{2}{*}{\shortstack{GPT-4o\\Flu.}}
			& \multirow{2}{*}{\shortstack{GPT-4o\\Ade.}} \\
			
			\cmidrule(lr){3-5}
			\cmidrule(lr){6-7}
			
			& & R-1 & R-2 & R-L & B-1 & B-4 & & & & \\
			
			\midrule
			
			\multirow{2}{*}{LLaMA3-8B}
			& Thought2Text
			& 30.0
			& 8.1
			& 26.6
			& 25.5
			& 5.5
			& 26.3
			& 0.890
			& \textbf{4.820}
			& 1.580 \\
			
			& \textbf{Ours}
			& \bestchg{35.4}{+5.4}
			& \bestchg{12.1}{+4.0}
			& \bestchg{31.9}{+5.3}
			& \bestchg{28.7}{+3.2}
			& \bestchg{8.1}{+2.6}
			& \bestchg{29.6}{+3.3}
			& \bestchg{0.898}{+0.008}
			& \valchg{4.745}{-0.075}
			& \bestchg{1.852}{+0.272} \\
			
			\midrule
			
			\multirow{2}{*}{Mistral-7B-v0.3}
			& Thought2Text
			& 30.6
			& 8.8
			& 28.0
			& 26.0
			& 6.1
			& 26.2
			& 0.890
			& \textbf{4.790}
			& 1.650 \\
			
			& \textbf{Ours}
			& \bestchg{36.2}{+5.6}
			& \bestchg{12.9}{+4.1}
			& \bestchg{32.9}{+4.9}
			& \bestchg{29.6}{+3.6}
			& \bestchg{8.5}{+2.4}
			& \bestchg{30.7}{+4.5}
			& \bestchg{0.900}{+0.010}
			& \valchg{4.776}{-0.014}
			& \bestchg{1.926}{+0.276} \\
			
			\midrule
			
			\multirow{4}{*}{Qwen2.5-7B}
			& Thought2Text
			& 26.4
			& 4.6
			& 22.8
			& 22.7
			& 3.7
			& 21.1
			& 0.880
			& 4.750
			& 1.280 \\
			
			& SENSE~\cite{murhekar2026sense}
			& 21.8
			& 5.9
			& 19.9
			& 13.8
			& 3.2
			& 15.0
			& 0.876
			& --
			& -- \\
			
			& SYNAPSE~\cite{murhekar2026synapse}
			& 32.2
			& 9.3
			& 28.7
			& \textbf{27.3}
			& 6.1
			& \textbf{28.0}
			& \textbf{0.904}
			& --
			& -- \\
			
			& \textbf{Ours}
			& \bestchg{33.6}{+1.4}
			& \bestchg{9.8}{+0.5}
			& \bestchg{29.7}{+1.0}
			& \bestchg{27.3}{+0.0}
			& \bestchg{6.5}{+0.4}
			& \valchg{27.2}{-0.8}
			& \valchg{0.894}{-0.010}
			& \textbf{4.780}
			& \textbf{1.873} \\
			
			\bottomrule
		\end{tabular}%
	}
	
\end{table}
\subsection{Subject-wise Stability}
As shown in Table~\ref{tab:subject_average_llm_results},  we further evaluate EEG-to-text generation
separately on the six available subjects under the all-subject
training setting. The proposed framework maintains relatively
consistent performance across subjects and LLM backbones.
The mean ROUGE-1 scores are $32.23 \pm 1.87$,
$27.58 \pm 1.56$, and $32.35 \pm 1.90$ for Mistral, Qwen,
and LLaMA, respectively. The moderate inter-subject standard
deviations indicate that the overall performance is not dominated
by a single participant, although subject variability remains
non-negligible. Detailed per-subject results are provided in the
supplementary material.
\begin{table}[!htbp]
	\centering
	\caption{
		Subject-wise EEG-to-text performance under the all-subject
		training setting. Values denote mean $\pm$ standard deviation
		across the six observed subjects and characterize inter-subject
		performance variability within the evaluated participant population.
	}
	\label{tab:subject_average_llm_results}
	\renewcommand{\arraystretch}{1.12}
	\setlength{\tabcolsep}{4.5pt}
	\resizebox{\textwidth}{!}{
		\begin{tabular}{lccccccc}
			\toprule
			\multirow{2}{*}{LLM Backbone}
			& \multicolumn{3}{c}{ROUGE}
			& \multicolumn{2}{c}{BLEU}
			& \multirow{2}{*}{METEOR}
			& \multirow{2}{*}{\makecell{BERT\\Score}} \\
			\cmidrule(lr){2-4} \cmidrule(lr){5-6}
			& R-1 & R-2 & R-L & B-1 & B-4 & & \\
			\midrule
			
			Mistral-7B-v0.3
			& 32.23 $\pm$ 1.87
			& \textbf{9.77 $\pm$ 1.57}
			& \textbf{29.07 $\pm$ 1.73}
			& \textbf{26.05 $\pm$ 1.07}
			& \textbf{6.27 $\pm$ 0.75}
			& \textbf{27.17 $\pm$ 2.23}
			& \textbf{0.894 $\pm$ 0.005} \\
			
			Qwen2.5-7B
			& 27.58 $\pm$ 1.56
			& 4.93 $\pm$ 0.42
			& 23.40 $\pm$ 1.14
			& 22.93 $\pm$ 1.43
			& 3.73 $\pm$ 0.23
			& 21.68 $\pm$ 1.15
			& 0.882 $\pm$ 0.002 \\
			
			LLaMA3-8B
			& \textbf{32.35 $\pm$ 1.90}
			& 9.45 $\pm$ 1.76
			& 28.90 $\pm$ 1.74
			& 25.95 $\pm$ 1.23
			& 6.07 $\pm$ 0.83
			& 26.38 $\pm$ 2.75
			& 0.893 $\pm$ 0.004 \\
			
			\bottomrule
		\end{tabular}
	}
\end{table}

\subsection{Cross-Backbone Improvement Visualization}
\begin{figure}[!htbp]
	\centering
	\includegraphics[width=0.7\textwidth,trim=52mm 22mm 18mm 10mm,clip]{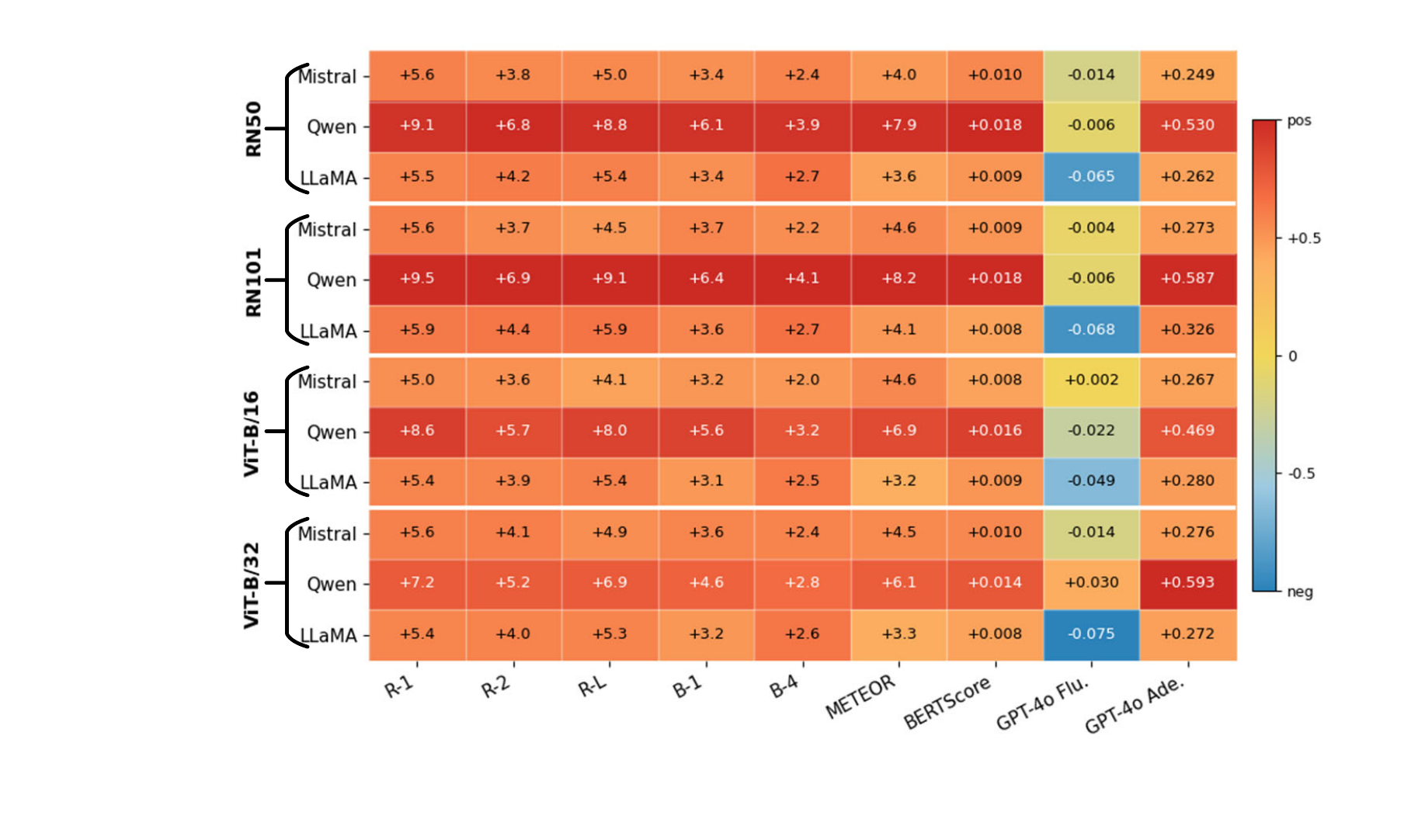}
	\caption{
		Metric-wise performance differences between our method and
		Thought2Text across different vision encoders and LLM backbones.
		Each cell reports the absolute score change with respect to the
		corresponding Thought2Text baseline under the same LLM backbone.
		Warm colors indicate performance gains, while blue colors indicate
		performance drops. 
	}
	\label{fig:delta_heatmap}
\end{figure}

Fig.~\ref{fig:delta_heatmap} visualizes the metric-wise
performance differences between our method and Thought2Text
across different vision encoders and LLM backbones.
The figure shows that our method achieves broadly positive gains
over Thought2Text across most settings, especially on the major
generation metrics such as ROUGE, BLEU, and METEOR.
This trend indicates that the proposed framework provides consistent
cross-backbone gains across the evaluated configurations, rather than
improvements restricted to a single model setting.

\subsection{Temporal Contribution Analysis}

As shown in Table~\ref{tab:temporal_masking_combined}, we further investigate how different
temporal portions of the EEG signal contribute to category
decoding and text generation.
The 440-sample EEG sequence is divided into five consecutive non-overlapping segments, and we perform inference-time keep/drop masking without retraining.
When individual segments are retained, the intermediate intervals $t_2$--$t_4$ consistently preserve more classification and generation performance than the boundary intervals $t_1$ and $t_5$.
In particular, $t_2$ achieves the strongest isolated Top-1 accuracy (26.0\%) and ROUGE-1 score (29.8).
Conversely, removing $t_2$ from the full EEG signal reduces Top-1 accuracy from 58.94\% to 44.7\% and ROUGE-1 from 36.2 to 33.8.
The consistent trends across Stage-1 classification and Stage-2 generation suggest that intermediate EEG responses provide useful semantic information for both category discrimination and downstream language generation.
\begin{table}[!htbp]
	\centering
	\caption{
		Combined temporal masking analysis for Stage-1 EEG classification
		and Stage-2 EEG-to-text generation. MSE-S and MSE-R denote the
		alignment errors to the sketch-image and raw-image CLIP embeddings,
		respectively. $\uparrow$ and $\downarrow$ indicate higher-is-better
		and lower-is-better metrics, respectively.
	}
	\label{tab:temporal_masking_combined}
	
	\renewcommand{\arraystretch}{0.92}
	\setlength{\tabcolsep}{2.8pt}
	\scriptsize
	
	\begin{adjustbox}{
			max width=\textwidth,
			max totalheight=0.76\textheight,
			keepaspectratio
		}
		\begin{tabular}{@{}lccccccccccc@{}}
			\toprule
			
			\multirow{2}{*}{\textbf{Setting}}
			&
			\multicolumn{4}{c}{\textbf{Stage-1 Classification}}
			&
			\multicolumn{7}{c}{\textbf{Stage-2 EEG-to-Text Generation}}
			\\
			
			\cmidrule(lr){2-5}
			\cmidrule(lr){6-12}
			
			&
			\textbf{Top-1$\uparrow$}
			&
			\textbf{Top-5$\uparrow$}
			&
			\textbf{MSE-S$\downarrow$}
			&
			\textbf{MSE-R$\downarrow$}
			&
			\textbf{R-1$\uparrow$}
			&
			\textbf{R-2$\uparrow$}
			&
			\textbf{R-L$\uparrow$}
			&
			\textbf{B-1$\uparrow$}
			&
			\textbf{B-4$\uparrow$}
			&
			\textbf{METEOR$\uparrow$}
			&
			\textbf{BERT$\uparrow$}
			\\
			
			\midrule
			
			\textbf{Full EEG}
			& \textbf{58.94}
			& \textbf{97.3}
			& \textbf{41.19}
			& \textbf{52.32}
			& \textbf{36.2}
			& \textbf{12.9}
			& \textbf{32.9}
			& \textbf{29.6}
			& \textbf{8.5}
			& \textbf{30.7}
			& \textbf{0.900}
			\\
			
			\midrule
			
			Keep $t_1$
			& 8.60 & 26.4 & 62.5 & 100.7
			& 27.6 & 4.80 & 24.5 & 23.2 & 4.00 & 21.3 & 0.882
			\\
			
			Keep $t_2$
			& 26.0 & 72.6 & 44.6 & 93.0
			& 29.8 & 7.40 & 26.6 & 24.6 & 5.10 & 23.8 & 0.887
			\\
			
			Keep $t_3$
			& 21.7 & 63.5 & 63.2 & 63.2
			& 29.7 & 6.90 & 26.6 & 24.4 & 4.80 & 23.3 & 0.885
			\\
			
			Keep $t_4$
			& 20.8 & 60.4 & 105.9 & 60.3
			& 29.5 & 6.6 & 26.2 & 24.5 & 4.60 & 23.1 & 0.884
			\\
			
			Keep $t_5$
			& 5.50 & 21.2 & 109.8 & 78.0
			& 26.5 & 5.50 & 23.6 & 22.1 & 3.80 & 20.3 & 0.880
			\\
			
			\midrule
			
			Drop $t_2$
			& 44.7 & 92.3 & 46.8 & 52.4
			& 33.8 & 10.5 & 30.6 & 27.6 & 6.80 & 28.0 & 0.894
			\\
			
			Drop $t_3$
			& 49.9 & 93.8 & 41.6 & 55.2
			& 34.4 & 11.5 & 31.3 & 28.1 & 7.70 & 28.7 & 0.896
			\\
			
			Drop $t_4$
			& 46.9 & 92.9 & 41.2 & 58.6
			& 34.2 & 11.2 & 31.1 & 28.1 & 7.50 & 28.7 & 0.895
			\\
			
			\bottomrule
		\end{tabular}
	\end{adjustbox}
\end{table}
\subsection{Generalization Across EEG Encoders}

To examine whether the proposed EEG-to-text framework depends on a specific neural backbone, we integrate several representative EEG encoders into the same generation pipeline.
As shown in Table~\ref{tab:eeg_backbone_generalization}, conventional EEG backbones also provide effective
EEG-conditioned generation under the same pipeline, while the
proposed dual-view encoder achieves the strongest or competitive
performance across most backbone--LLM combinations.
Under Mistral-7B-v0.3, our encoder improves ROUGE-1 from 33.9 to 36.2 over the strongest conventional backbone and also obtains higher METEOR and BERTScore.
These results support the contribution of dual-view semantic EEG representation beyond the downstream language-generation pipeline.
\begin{table}[!htbp]
	\centering
	\caption{
		Generalization of the proposed EEG-to-text training framework across different EEG backbones.
		All EEG backbones are trained and evaluated under the same EEG-to-text training pipeline.
		\textit{Ours} denotes the proposed dual-view EEG semantic token encoder.
		The best value for each metric within each LLM group is highlighted in bold.
	}
	\label{tab:eeg_backbone_generalization}
	
	\renewcommand{\arraystretch}{1.02}
	\setlength{\tabcolsep}{3.8pt}
	\footnotesize
	
	\resizebox{\textwidth}{!}{%
		\begin{tabular}{llccccccccc}
			\toprule
			
			\multirow{2}{*}{LLM}
			& \multirow{2}{*}{EEG Backbone}
			& \multicolumn{3}{c}{ROUGE}
			& \multicolumn{2}{c}{BLEU}
			& \multirow{2}{*}{METEOR}
			& \multirow{2}{*}{\makecell{BERT\\Score}}
			& \multirow{2}{*}{\makecell{GPT-4o\\Flu.}}
			& \multirow{2}{*}{\makecell{GPT-4o\\Ade.}} \\
			
			\cmidrule(lr){3-5}
			\cmidrule(lr){6-7}
			
			& & R-1 & R-2 & R-L & B-1 & B-4 & & & & \\
			
			\midrule
			
			\multirow{5}{*}{LLaMA3-8B}
			& EEGNet
			& 29.1 & 6.6 & 26.0 & 24.0 & 4.8 & 22.3 & 0.886 & 4.771 & 1.311 \\
			& ATCNet
			& 30.2 & 7.8 & 27.1 & 24.5 & 5.4 & 23.4 & 0.888 & 4.768 & 1.418 \\
			& EEGConformer
			& 33.4 & 10.6 & 30.0 & 27.1 & 7.1 & 27.3 & 0.895 & 4.745 & 1.700 \\
			& ShallowConvNet
			& 33.6 & 10.6 & 30.2 & 27.8 & 7.1 & 27.6 & 0.895 & \textbf{4.772} & 1.701 \\
			& Ours
			& \textbf{35.4} & \textbf{12.1} & \textbf{31.9}
			& \textbf{28.7} & \textbf{8.1} & \textbf{29.6}
			& \textbf{0.898} & 4.745 & \textbf{1.852} \\
			
			\midrule
			
			\multirow{5}{*}{Mistral-7B-v0.3}
			& EEGNet
			& 29.8 & 7.1 & 26.5 & 24.7 & 4.8 & 23.5 & 0.886 & \textbf{4.790} & 1.359 \\
			& ATCNet
			& 30.4 & 7.8 & 27.4 & 24.9 & 5.3 & 24.5 & 0.889 & 4.787 & 1.437 \\
			& EEGConformer
			& 33.9 & 11.2 & 30.7 & 27.5 & 7.4 & 28.2 & 0.896 & 4.722 & 1.804 \\
			& ShallowConvNet
			& 33.2 & 10.1 & 29.8 & 27.2 & 6.8 & 27.5 & 0.894 & 4.772 & 1.720 \\
			& Ours
			& \textbf{36.2} & \textbf{12.9} & \textbf{32.9}
			& \textbf{29.6} & \textbf{8.5} & \textbf{30.7}
			& \textbf{0.900} & 4.776 & \textbf{1.926} \\
			
			\midrule
			
			\multirow{5}{*}{Qwen2.5-7B}
			& EEGNet
			& 29.7 & 6.2 & 26.0 & 24.4 & 4.5 & 22.2 & 0.885 & 4.752 & 1.300 \\
			& ATCNet
			& 30.4 & 6.6 & 26.6 & 24.8 & 4.8 & 22.8 & 0.888 & 4.729 & 1.364 \\
			& EEGConformer
			& \textbf{33.7} & \textbf{10.0} & \textbf{30.0}
			& 26.9 & \textbf{6.7} & 27.5 & 0.895 & 4.700 & 1.689 \\
			& ShallowConvNet
			& 31.5 & 7.8 & 27.6 & 26.3 & 5.4 & 24.9 & 0.891 & 4.720 & 1.526 \\
			& Ours
			& 33.6 & 9.8 & 29.7
			& \textbf{27.3} & 6.5 & 27.2
			& 0.894 & \textbf{4.780} & \textbf{1.873} \\
			
			\bottomrule
		\end{tabular}%
	}
\end{table}

\section{Ablation Studies}

\label{sec:ablation}
\subsection{Component Ablation}
\label{sec:ablation_components}

To assess the contribution of each component, we conduct ablations under identical backbones, training schedules, and decoding settings. As shown in Table~\ref{tab:ablation_eeg2text}, the full model achieves the strongest overall performance, while removing diversity regularization consistently degrades text-generation metrics, indicating that it helps prevent EEG-token collapse. All dual-view variants, including single-view input, view removal, view swapping, and duplicated views, reduce performance, supporting the complementarity of the raw-oriented and
sketch-oriented EEG representations. The larger drops observed for Shared avg. and Direct projector show
that average-pooled representations or unconstrained EEG-to-LLM
mapping provide a weaker language interface than branch-specific
semantic querying. Stage-wise results reveal that Stages A and B mainly stabilize visual-semantic initialization and cross-view adaptation, whereas removing Stage C causes the largest degradation, demonstrating the importance of EEG-conditioned language adaptation. By contrast, the Stage-B objectives and auxiliary text-semantic alignment provide
modest improvements on most metrics, suggesting auxiliary roles in
optimization stability and semantic compatibility. Overall, the improvements primarily arise from combining dual-view EEG semantic tokenization with EEG-conditioned language adaptation.
\begin{table}[!htbp]
	\centering
	\caption{Ablation study of the proposed EEG-to-text fine-tuning framework. }
	\label{tab:ablation_eeg2text}
	\renewcommand{\arraystretch}{1.12}
	\setlength{\tabcolsep}{3.2pt}
	\resizebox{\textwidth}{!}{
		\begin{tabular}{llccccccc|cc}
			\toprule
			\multirow{2}{*}{Variant}
			& \multirow{2}{*}{Modification}
			& \multicolumn{3}{c}{ROUGE}
			& \multicolumn{2}{c}{BLEU}
			& \multirow{2}{*}{METEOR}
			& \multirow{2}{*}{\makecell{BERT\\Score}}
			& \multirow{2}{*}{\makecell{GPT-4o\\Flu.}}
			& \multirow{2}{*}{\makecell{GPT-4o\\Ade.}} \\
			\cmidrule(lr){3-5} \cmidrule(lr){6-7}
			& & R-1 & R-2 & R-L & B-1 & B-4 & & & & \\
			\midrule
			
			\multicolumn{11}{l}{\textit{Full setting and regularization ablation}} \\
			\midrule
			Full model
			& \makecell[l]{Full setting}
			& \textbf{36.2} & \textbf{12.9} & \textbf{32.9} & \textbf{29.6}& \textbf{8.5} & \textbf{30.7}  & \textbf{0.900} & \textbf{4.776} & \textbf{1.926}  \\
			
			w/o Div.
			& \makecell[l]{Remove diversity regularization}
			& 34.0 & 11.4 & 31.4 & 27.1 & 7.5 & 28.1 & 0.899 & 4.761 & 1.833 \\
			
			\midrule
			\multicolumn{11}{l}{\textit{Dual-view input ablations}} \\
			\midrule
			Raw only, sketch zeroed
			& \makecell[l]{Use raw-view EEG while zeroing the sketch-view input}
			& 34.9 & 11.5 & 31.6 & 28.2 & 7.5 & 29.4 & 0.897 & 4.741  & 1.869 \\
			
			Sketch only, raw zeroed
			& \makecell[l]{Use sketch-view EEG while zeroing the raw-view input}
			& 33.4 & 10.2 & 30.1 & 26.8 & 6.7 & 27.7 & 0.896 & 4.765  & 1.776 \\
			
			Swap raw/sketch
			& \makecell[l]{Swap the raw-view and sketch-view EEG branches}
			& 35.3 & 12.1 & 32.1 & 28.6 & 7.9 & 29.8 & \textbf{0.900} & 4.775 & 1.899 \\
			
			Sketch as both views
			& \makecell[l]{Feed the sketch-view representation to both branches}
			& 35.8 & 12.4 & 32.6 & 29.0 & 8.2 & 30.2 & \textbf{0.900} & 4.775  & 1.899 \\
			
			Raw as both views
			& \makecell[l]{Feed the raw-view representation to both branches}
			& 35.8 & 12.4 & 32.6 & 29.3 & 8.1 & 30.4 & 0.899 & 4.747  & 1.896 \\
			
			Raw only
			& \makecell[l]{Remove the sketch-view branch and use raw-view EEG only}
			& 35.4 & 12.4 & 32.3 & 28.6 & 8.1 & 29.8 & 0.899 & 4.766 & 1.900 \\
			
			Sketch only
			& \makecell[l]{Remove the raw-view branch and use sketch-view EEG only}
			& 35.2 & 11.7 & 31.7 & 28.8 & 7.7 & 29.6 & 0.898 & 4.768 & 1.873 \\
			Shared avg.
			& \makecell[l]{Use shared average pooling instead of branch-specific semantic slots}
			& 33.3 & 9.4 & 29.5 & 27.1 & 6.3 & 26.8 & 0.893 & 4.765 & 1.916 \\
			\midrule
			\multicolumn{11}{l}{\textit{Semantic query bridge ablations}} \\
			\midrule
			Direct projector
			& Replace the Stage-C bridge with a direct EEG-to-LLM projection
			& 33.4
			& 8.8
			& 29.1
			& 26.4
			& 6.0
			& 27.2
			& 0.894
			& 4.766
			& 1.916 \\
			
			Query projector
			& Retain query pooling and text projection only
			& 35.6
			& 12.4
			& 32.4
			& 29.1
			& 8.2
			& 30.3
			& 0.898
			& 4.752
			& 1.910 \\
			
			w/o Semantic Align.
			& Retain the residual bridge but remove branch text-semantic alignment
			& 35.8 & 12.5
			& 32.5
			& 29.1
			& 8.3
			& 30.2
			& \textbf{0.900}
			& 4.762
			& 1.922 \\
			
			\midrule
			\multicolumn{11}{l}{\textit{Stage-wise training ablations}} \\
			\midrule
			w/o Stage A
			& \makecell[l]{Skip the Stage-A warm-up procedure}
			& 34.6 & 11.6 & 31.2 & 28.6 & 7.7 & 30.2 & 0.896 & 4.764 & 1.876 \\
			
			w/o Stage B
			& \makecell[l]{Skip the Stage-B intermediate alignment procedure}
			& 35.7 & 12.3 & 32.5 & 28.7 & 8.1 & 29.6 & 0.899 & 4.754 & 1.899 \\
			
			w/o Stage C
			& \makecell[l]{Skip the Stage-C EEG-conditioned language fine-tuning}
			& 17.2 & 2.8 & 14.6 & 14.7 & 2.8 & 18.3 & 0.874 & 4.572 & 1.749 \\
			
			\bottomrule
		\end{tabular}
	}
\end{table}
\subsection{Qualitative Analysis of Dual-View Complementarity}

To complement the branch-level representation analysis, we further examine how dual-view supervision affects the generated descriptions.
Table~\ref{tab:qualitative_view_ablation} compares the full raw+sketch model with raw-only and sketch-only variants on representative examples containing fine-grained visual attributes.
The dual-view model more consistently preserves specific cues that are missed or altered by the single-view variants, including the yellow cap and brown gills of the mushroom, the black strap of the wristwatch, the wooden table in the camera scene, and the blue stripe of the sneakers.
These examples qualitatively support the complementarity of raw- and sketch-oriented EEG representations and illustrate how their combination can improve fine-grained semantic preservation in EEG-conditioned text generation.
\begin{table}[h]
	\centering
	\caption{
		Qualitative comparison among raw+sketch, raw-only, and sketch-only
		supervision. The examples emphasize color, texture, material, and
		part attributes. Blue text highlights representative visual cues,
		green text emphasizes descriptions produced by the dual-view model,
		and red text marks incorrect attributes generated by single-view variants.
	}
	\label{tab:qualitative_view_ablation}
	
	\renewcommand{\arraystretch}{1.15}
	\setlength{\tabcolsep}{2.2pt}
	\scriptsize
	
	\begin{tabularx}{\textwidth}{
			C{0.035\textwidth}
			C{0.115\textwidth}
			C{0.125\textwidth}
			Y
			Y
			Y
			Y
		}
		
		\toprule
		
		\textbf{ID}
		&
		\textbf{Image}
		&
		\makecell[c]{\textbf{Reference}\\\textbf{Object}}
		&
		\makecell[c]{\textbf{Reference}\\\textbf{Description}}
		&
		\makecell[c]{\textbf{Raw + Sketch}\\\textbf{Description}}
		&
		\makecell[c]{\textbf{Raw-only}\\\textbf{Description}}
		&
		\makecell[c]{\textbf{Sketch-only}\\\textbf{Description}}
		\\
		
		\midrule
		
		1
		&
		\QualImg{Figures/image_dataset/mushroom.jpg}
		&
		mushroom
		&
		A large \cue{yellow} mushroom with a \cue{brown} stem and a
		\cue{brown} cap, surrounded by green foliage.
		&
		\good{A close-up of a mushroom with a \cue{yellow cap} and
			\cue{brown gills}.}
		&
		A mushroom growing out of a log.
		&
		A mushroom with a \cue{yellow} background.
		\\
		
		\midrule
		
		2
		&
		\QualImg{Figures/image_dataset/n03197337_4120.JPEG}
		&
		\makecell[c]{watch or\\watches}
		&
		A \cue{silver} wristwatch with a digital display and a
		\cue{black strap}.
		&
		\good{A \cue{silver wristwatch} with a \cue{black strap}
			and a digital display.}
		&
		A collection of watches on display.
		&
		A black watch with a white face and red numbers.
		\\
		
		\midrule
		
		3
		&
		\QualImg{Figures/image_dataset/n04069434_9580.JPEG}
		&
		camera
		&
		A vintage camera with a lens on a \cue{wooden table}.
		&
		\good{A vintage black and white camera with a lens on a
			\cue{wooden table}.}
		&
		A black camera with a silver lens and strap.
		&
		A vintage camera with a leather strap.
		\\
		
		\midrule
		
		4
		&
		\QualImg{Figures/image_dataset/n04120489_6792.JPEG}
		&
		shoes
		&
		A pair of \cue{white and blue sneakers}.
		&
		\good{A pair of white and black sneakers with a
			\cue{blue stripe}.}
		&
		A pair of white and black sneakers with a \bad{red tag}.
		&
		A pair of white and black sneakers with a \bad{red sole}.
		\\
		
		\bottomrule
	\end{tabularx}
	
\end{table}

\subsection{Branch-Level View Specialization Analysis}
The downstream ablations in Table~\ref{tab:ablation_eeg2text} indicate that the two EEG branches are complementary, but do not directly establish whether they learn view-specific representations. We therefore perform a branch-level cross-view alignment analysis on all 1,987 test samples. For each sample, we compute the cosine similarity between the sketch- and raw-oriented EEG representations and both sketch- and raw-image CLIP embeddings. We define the specialization margin as the similarity to the matched visual view minus that to the alternative view. As shown in Table~\ref{tab:branch_specialization}, the sketch branch exhibits a strong preference for the sketch view, with a specialization margin of 0.1900 (95\% CI: [0.1871, 0.1929]). The raw branch shows a smaller but consistently positive preference for the raw view, with a margin of 0.0195 (95\% CI: [0.0168, 0.0223]). The confidence intervals are estimated using 5,000 paired-bootstrap resamples. These results provide direct representational evidence that the two EEG branches do not learn identical mappings, but exhibit different degrees of view-specific specialization.

The specialization trend is also consistent at the sample level:
99.85\% of test samples show a positive sketch-view preference,
while 61.40\% show a positive raw-view preference, with both
preferences simultaneously positive for 61.25\% of samples.
At the category level, the mean sketch specialization margin is
positive for all 40 categories, whereas the raw-view preference is
positive for 25 categories and is more category dependent.
These results indicate that sketch-view specialization is consistently
observed across the evaluated test samples and categories, whereas
raw-view specialization is weaker and more category dependent.

Aggregating both branches, the mean matched-view similarity is
0.7804 compared with 0.6756 for the crossed assignment,
yielding an overall diagonal advantage of 0.1048
(95\% CI: [0.1033, 0.1062]).

\begin{table}[!htbp]
	\centering
	\caption{
		Branch-level cross-view alignment on 1,987 test samples.
		Values are mean cosine similarities; the margin is the matched-view
		minus alternative-view similarity. Brackets denote 95\% CIs from
		5,000 paired-bootstrap resamples.
	}
	\label{tab:branch_specialization}
	
	\footnotesize
	\setlength{\tabcolsep}{3pt}
	\renewcommand{\arraystretch}{1.05}
	
	\begin{tabular}{lccc}
		\toprule
		Branch &
		Sketch &
		Raw &
		Margin (95\% CI) \\
		\midrule
		
		Sketch &
		\textbf{0.8300} &
		0.6400 &
		\textbf{+0.1900 [0.1871, 0.1929]} \\
		
		Raw &
		0.7113 &
		\textbf{0.7308} &
		\textbf{+0.0195 [0.0168, 0.0223]} \\
		
		\bottomrule
	\end{tabular}
\end{table}

\subsection{EEG-Label Disentanglement}

To determine whether generation depends on EEG-derived soft
prompts rather than object-label prompts or language priors,
we perform inference-time interventions without retraining.
As shown in Table~\ref{tab:inference_sanity_checks}, globally shuffling or zeroing the EEG
prompts reduces ROUGE-1 from 36.2 to 23.9 and 21.9,
respectively, demonstrating the importance of EEG-derived
information.

However, global shuffling simultaneously disrupts category
consistency and sample-specific EEG-image correspondence.
To isolate sample-level EEG information beyond category identity,
we therefore introduce a within-category EEG shuffle, in which
EEG soft prompts are exchanged only among samples sharing the
same object category. This intervention preserves category
identity while breaking the original EEG-image correspondence
and reduces ROUGE-1 from 36.2 to 32.6, indicating that
EEG-derived soft prompts contain information beyond
object-category identity.

Label prompts alone are insufficient: the ``Predicted label only''
and ``Oracle label only'' variants achieve ROUGE-1 scores of
16.9 and 18.5, respectively, whereas removing the label prompt
retains 32.0 ROUGE-1. Finally, the EEG + oracle-label diagnostic configuration achieves the
highest scores, showing that accurate category information can complement
EEG-derived sample-specific information. Because this setting uses
ground-truth category labels at inference, it should be interpreted as an
oracle analysis rather than a deployable configuration.
\begin{table}[!htbp]
	\centering
	\caption{Inference-stage EEG--label disentanglement analysis.}
	\label{tab:inference_sanity_checks}
	
	\renewcommand{\arraystretch}{1.10}
	\setlength{\tabcolsep}{3.4pt}
	\scriptsize
	
	\begin{adjustbox}{max width=\textwidth}
		\begin{tabular}{
				@{}
				>{\raggedright\arraybackslash}m{0.130\textwidth}
				>{\raggedright\arraybackslash}m{0.235\textwidth}
				ccccc
				c
				c
				cc
				@{}
			}
			
			\toprule
			
			\multirow{2}{*}{\textbf{Variant}}
			&
			\multirow{2}{*}{\textbf{Modification}}
			&
			\multicolumn{3}{c}{\textbf{ROUGE}}
			&
			\multicolumn{2}{c}{\textbf{BLEU}}
			&
			\multirow{2}{*}{\textbf{METEOR}}
			&
			\multirow{2}{*}{\makecell[c]{\textbf{BERT}\\\textbf{Score}}}
			&
			\multicolumn{2}{c}{\textbf{GPT-4o}}
			\\
			
			\cmidrule(lr){3-5}
			\cmidrule(lr){6-7}
			\cmidrule(lr){10-11}
			
			&
			&
			\textbf{R-1}
			&
			\textbf{R-2}
			&
			\textbf{R-L}
			&
			\textbf{B-1}
			&
			\textbf{B-4}
			&
			&
			&
			\textbf{Flu.}
			&
			\textbf{Ade.}
			\\
			
			\midrule
			
			\textbf{Full model}
			&
			EEG soft prompts + predicted label prompt
			&
			36.2
			& 12.9
			& 32.9
			& 29.6
			& 8.5
			& 30.7
			& 0.900
			& 4.776
			& 1.926
			\\
			
			\midrule
			
			\makecell[l]{Within\\-category \\ shuffle}
			&
			Shuffle EEG soft prompts within the same predicted category
			&
			32.6
			& 11.0
			& 29.3
			& 26.7
			& 7.9
			& 29.3
			& 0.899
			& 4.754
			& 1.850
			\\[2pt]
			\midrule
			Shuffle EEG
			&
			Shuffle EEG soft prompts across test samples
			&
			23.9
			& 4.1
			& 21.9
			& 19.5
			& 3.3
			& 17.8
			& 0.877
			& 4.733
			& 1.048
			\\[2pt]
			\midrule
			Zero EEG
			&
			Set EEG soft prompts to zero
			&
			21.9
			& 4.6
			& 19.0
			& 17.8
			& 3.8
			& 21.7
			& 0.883
			& 4.656
			& 1.633
			\\[2pt]
			\midrule
			\makecell[l]{No label\\prompt}
			&
			Remove the predicted object-label prompt
			&
			32.0
			& 8.9
			& 28.7
			& 26.2
			& 6.1
			& 26.2
			& 0.890
			& 4.739
			& 1.529
			\\[2pt]
			\midrule
			\makecell[l]{Zero EEG \\ +no label}
			&
			Remove both EEG and label prompts
			&
			18.1
			& 0.7
			& 16.9
			& 15.2
			& 2.3
			& 15.4
			& 0.867
			& 4.653
			& 1.053
			\\[2pt]
			\midrule
			\makecell[l]{Pred. label\\only}
			&
			Remove EEG; use predicted label only
			&
			16.9
			& 2.4
			& 14.2
			& 14.7
			& 2.4
			& 17.5
			& 0.863
			& 4.259
			& 1.670
			\\
			
			\midrule
			
			\makecell[l]{Oracle label\\only}
			&
			Remove EEG; use ground-truth label only
			&
			18.5
			& 3.1
			& 15.5
			& 16.0
			& 2.7
			& 20.0
			& 0.869
			& 4.359
			& 2.130
			\\[2pt]
			\midrule
			\textbf{\makecell[l]{EEG \\ + oracle\\label}}
			&
			EEG soft prompts + ground-truth label prompt
			&
			\textbf{42.8}
			& \textbf{17.7}
			& \textbf{38.8}
			& \textbf{34.8}
			& \textbf{11.4}
			& \textbf{38.1}
			& \textbf{0.912}
			& \textbf{4.777}
			& \textbf{2.501}
			\\
			
			\bottomrule
			
		\end{tabular}
	\end{adjustbox}
\end{table}
\FloatBarrier
\section{Conclusion}
\label{sec:conclusion}
In this paper, we presented a dual-view semantic token bridging framework for visually evoked EEG-to-text generation. EEG tokens learned under raw-image and sketch-image supervision
capture complementary view-oriented semantic representations,
while a branch-specific semantic query bridge transforms these
structured EEG representations into LLM-compatible soft prompts. Experiments and ablations show strong subject-wise EEG classification
performance and consistent EEG-to-text improvements over the
Thought2Text baseline across the evaluated LLM backbones. Limitations include the multi-stage training cost, evaluation
on a single EEG dataset with a limited participant population,
and the lack of validation across broader recording conditions,
electrode configurations, and open-category settings.
Future work will investigate broader datasets and more diverse
neural recording conditions.

\bibliographystyle{elsarticle-num}
\bibliography{reference}

\end{document}